\documentclass{article}
\usepackage[T1]{fontenc}
\usepackage{colm2024_conference}
\usepackage{booktabs}
\usepackage[table]{xcolor}
\definecolor{ResultInk}{HTML}{343057}
\definecolor{ResultHeader}{HTML}{E5DFF3}
\definecolor{ResultStripe}{HTML}{F7F7FA}
\definecolor{ResultHighlight}{HTML}{F0EBFA}
\definecolor{ResultTeal}{HTML}{227F80}

\usepackage{amsmath}
\usepackage{amsthm}
\theoremstyle{definition}

\usepackage{graphicx}
\usepackage{etoolbox}
\AtBeginEnvironment{figure}{\setlength{\abovecaptionskip}{-4pt}}
\AtBeginEnvironment{table}{%
  \setlength{\abovecaptionskip}{0pt}%
  \setlength{\belowcaptionskip}{5pt}%
}
\fancypagestyle{firstpage}{%
  \fancyhf{}
  \lhead{\raisebox{-2pt}{\includegraphics[height=18pt]{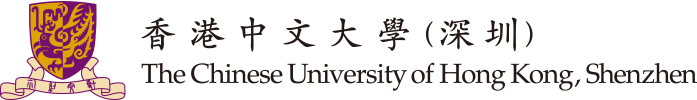}}}
  \rhead{\raisebox{-2pt}{\includegraphics[height=18pt]{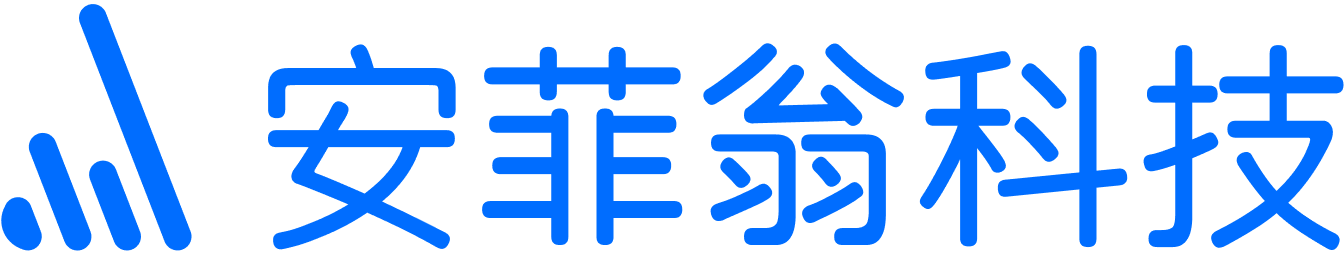}}}
}
\usepackage{array}
\usepackage{tabularx}
\usepackage{ragged2e}
\usepackage{placeins}
\usepackage{float}
\usepackage{pifont}
\usepackage[most]{tcolorbox}
\newtcolorbox{promptbox}[1]{%
  enhanced,breakable,width=\linewidth,
  colback=ResultStripe,colframe=ResultHeader,colbacktitle=ResultHeader,
  coltitle=ResultInk,boxrule=0.6pt,arc=2pt,
  left=8pt,right=8pt,top=6pt,bottom=6pt,
  toptitle=5pt,bottomtitle=5pt,
  fonttitle=\bfseries\small,fontupper=\small,
  before skip=10pt,after skip=10pt,
  title={#1}%
}
\newcommand{\splithdr}[2]{\begin{tabular}[t]{@{}c@{}}#1\\#2\end{tabular}}
\newcommand{\yes}{\ding{51}}
\newcommand{\no}{\ding{55}}

\newcolumntype{L}{>{\RaggedRight\arraybackslash}X}
\newcolumntype{Y}{>{\RaggedRight\arraybackslash}p}
\makeatletter
\def\CT@firstcol{0}
\def\CT@lastcol{0}
\def\CT@colmark{0}
\def\CT@colstack{}
\newif\ifCT@preambledone
\protected\def\CT@trim@overhang#1{%
  \ifnum#1=\CT@firstcol\relax \@tempdimb\z@ \fi
  \ifnum#1=\CT@lastcol\relax \@tempdimc\z@ \fi
}
\def\CT@recordcol{%
  \ifnum\CT@colmark=0
    \xdef\CT@firstcol{\the\count@}%
    \gdef\CT@colmark{1}%
  \fi
  \xdef\CT@lastcol{\the\count@}%
}
\def\CT@popcol#1#2#3\relax{%
  \gdef\CT@firstcol{#1}%
  \gdef\CT@lastcol{#2}%
  \gdef\CT@colstack{#3}%
}
\patchcmd{\@classz}{\columncolor!\@nil}{%
  \columncolor!\@nil\CT@recordcol
}{}{\PackageWarning{duplex}{Could not record table columns}}
\patchcmd{\@classz}{\CT@cell@color\CT@do@color}{%
  \CT@cell@color
  \noexpand\CT@trim@overhang{\the\count@}%
  \CT@do@color
}{}{\PackageWarning{duplex}{Could not patch column color overhang}}
\let\CT@mkpreamorig\@mkpream
\def\@mkpream#1{%
  \ifCT@preambledone
    \xdef\CT@colstack{{\CT@firstcol}{\CT@lastcol}\CT@colstack}%
  \fi
  \gdef\CT@colmark{0}%
  \CT@mkpreamorig{#1}%
  \ifCT@preambledone
    \expandafter\CT@popcol\CT@colstack\relax
  \else
    \global\CT@preambledonetrue
  \fi
}
\pretocmd{\CT@start}{%
  \xdef\CT@colstack{{\CT@firstcol}{\CT@lastcol}\CT@colstack}%
  \global\CT@preambledonefalse
}{}{}
\apptocmd{\CT@end}{%
  \expandafter\CT@popcol\CT@colstack\relax
}{}{}
\makeatother

\title{DuplexAgent-RSI: Recursive Harness Improvement for \\ Full-Duplex Voice Agent Collaboration}

\newcommand{\name}[1]{\def\reportauthors{#1}}
\newcommand{\address}[1]{\def\reportaddress{#1}}
\name{\begin{tabular}[t]{@{}c@{}}
Yingda Shen$^{\star}$ \hspace{1em} Yuxiang Wang$^{\star}$ \hspace{1em} Kunyu Feng$^{\star}$ \hspace{1em} Qinke Ni$^{\star}$ \hspace{1em} Jiaqi Li$^{\star\dagger}$\\
Minghao Hsu$^{\star}$ \hspace{1em} Junan Zhang$^{\star}$ \hspace{1em} Dekun Chen$^{\star}$ \hspace{1em} Yutong Bian$^{\star\dagger}$ \hspace{1em} Zhizheng Wu$^{\star\dagger}$
\end{tabular}}
\address{$^{\star}$ The Chinese University of Hong Kong, Shenzhen \\
$^{\dagger}$ Amphion Technology Co., Ltd}
\newcommand{\projecturl}{https://amphionteam.github.io/DuplexAgent-RSI-Demo/}
\newcommand{\githuburl}{https://amphionteam.github.io/DuplexAgent-RSI-Demo/}
\newcommand{\browserglobe}{%
  \textcolor{darkblue}{%
    \begin{tikzpicture}[baseline=(current bounding box.south),x=0.85em,y=0.85em,line cap=round]
      \draw[darkblue,line width=0.06em] (0,0) circle (0.42);
      \draw[darkblue,line width=0.05em] (0,0) ellipse (0.17 and 0.42);
      \draw[darkblue,line width=0.05em] (-0.42,0) -- (0.42,0);
      \draw[darkblue,line width=0.045em] (-0.36,0.20) .. controls (-0.12,0.28) and (0.12,0.28) .. (0.36,0.20);
      \draw[darkblue,line width=0.045em] (-0.36,-0.20) .. controls (-0.12,-0.28) and (0.12,-0.28) .. (0.36,-0.20);
    \end{tikzpicture}}%
}
\newcommand{\githublogo}{\raisebox{-0.16em}{\includegraphics[height=1.05em]{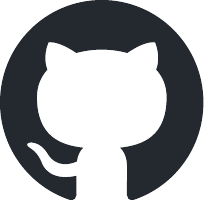}}}
\author{\reportauthors\\[3pt]\small\reportaddress}
\makeatletter
\def\@maketitle{%
  \vbox{\hsize\textwidth
    {\centering{\Large\bf \@title\par}}%
    \vspace{-0.3cm}%
    \begin{quote}\rule{\z@}{6pt}{\centering\par{\@author}\par}\end{quote}%
    \vspace{-0.55em}%
    {\centering\normalsize
      \href{\projecturl}{\browserglobe\ Project Page}\hspace{1.6em}%
      \href{\githuburl}{\githublogo\ GitHub}\par}%
    \vspace{0.15em}%
  }%
  \thispagestyle{firstpage}%
}
\makeatother
\renewenvironment{abstract}{%
  \par\vspace{0.1em}%
  {\centering\large\bf Abstract\par}%
  \vspace{0.15em}%
  \begingroup
  \setlength{\topsep}{0pt}%
  \setlength{\partopsep}{0pt}%
  \setlength{\parskip}{0pt}%
  \begin{quote}%
}{%
  \end{quote}%
  \endgroup
  \vspace{-0.15em}%
}

\begin{document}
\maketitle

\begin{abstract}
Voice agents are converging on a collaboration pattern: a full-duplex
interaction model stays on the live channel as the entry to the
conversation, while search, reasoning, and coding are handled through
asynchronous delegation. A duplex model supports continuous listening
and speaking, but complex reasoning and tool use may exceed its
capabilities. A coding agent can plan and execute extended tasks, but
its sequential interface is a poor fit for live conversation. Combining
them requires a harness that coordinates task acceptance, progress,
cancellation, replacement, and result delivery while keeping the
conversation responsive. Existing harnesses often rely on coupled
heuristics, making them difficult to improve systematically from
evidence. We present DuplexAgent, a full-duplex collaboration system
whose harness expresses this workflow as six editable modules, and
\emph{Duplex-Harness-RSI}, a closed loop that revises them from
interaction traces. A simulator automatically generates timed test
conversations, runs the system, and produces failure traces that
identify the collaboration modules requiring repair. Reasoning LLMs
and coding agents in the delegation pool also serve the improvement
loop: the Exam Planner selects the next tests from observed weaknesses
and the repair archive, and the Harness Editor proposes targeted module
changes. The capabilities that serve the user thus also improve the
system's coordination. Experiments on intelligence, agentic, and
duplex benchmarks show that DuplexAgent combines continuous
interaction with difficult reasoning and complex task execution,
achieving stronger spoken-knowledge and executable-tool scores than
the compared delegated systems while maintaining strong interruption
response. A harness ablation further shows that this modular,
verifiable loop outperforms the initial harness and repeated editing
that lacks its diagnosis and repair archive.
\end{abstract}



\setlength{\intextsep}{2pt plus 1pt minus 1pt}
\begin{figure}[!ht]
\centering
\includegraphics[width=\linewidth]{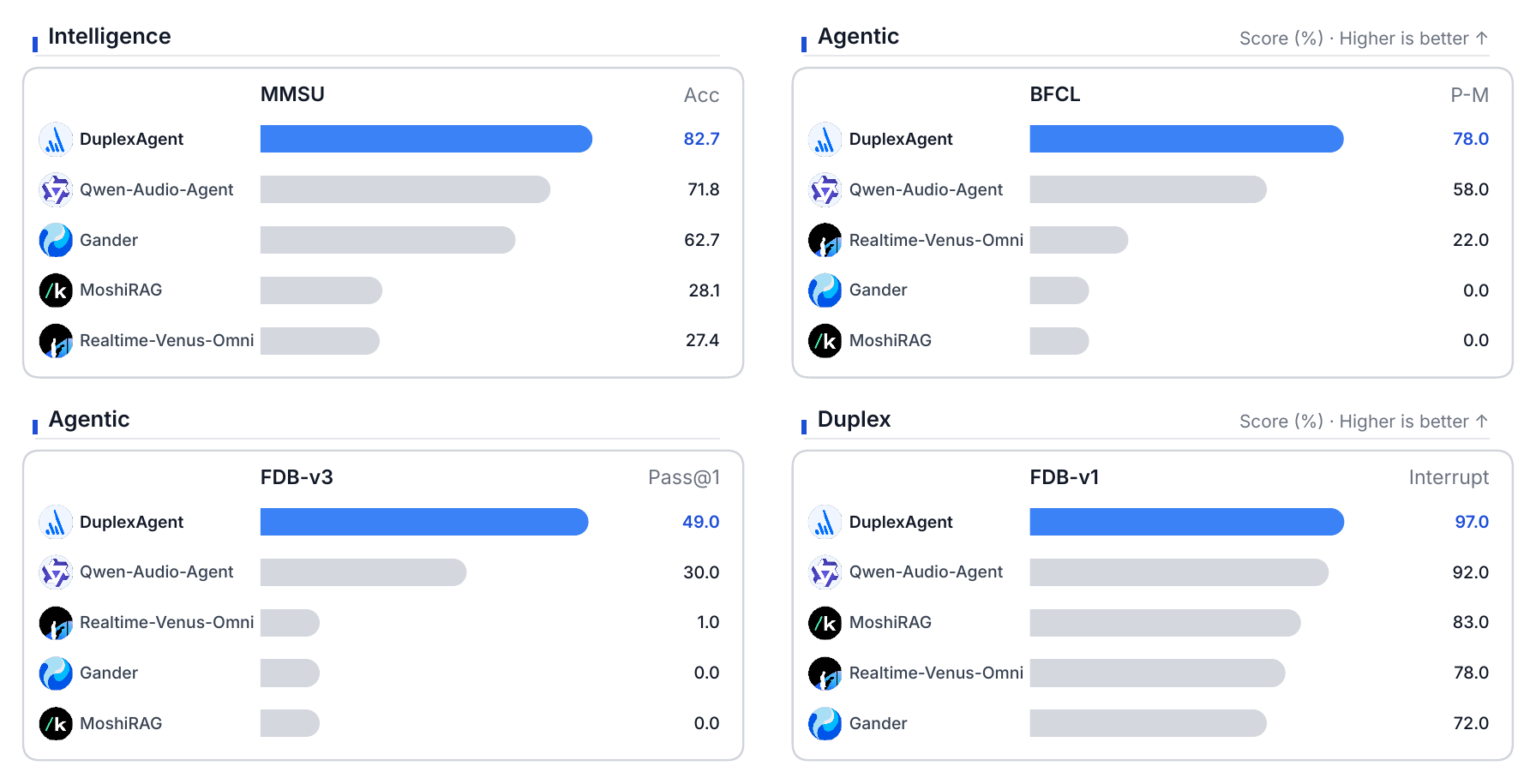}
\caption{Benchmark results for DuplexAgent (blue), compared with other collaboration systems (gray). Higher is better on every panel. Intelligence reports MMSU accuracy, agentic tool use reports BFCL parallel-multiple and FDB-v3 Pass@1, and duplex reports FDB-v1 interrupt TOR. DuplexAgent is stronger on all three capabilities. All collaboration systems use the same delegation LLM. See Tables~\ref{tab:agentic} and \ref{tab:intel-duplex}.}
\label{fig:leaderboard}
\end{figure}
\FloatBarrier
\setlength{\intextsep}{12pt plus 2pt minus 2pt}

\section{Introduction}
\label{sec:intro}

As language-model agents move from answering questions to carrying
out work such as search, tool use, multi-step plans, and code
edits~\citep{yao2023react}, users need to supervise that work as it
proceeds: they refine the goal, ask for status, interrupt an
answer, or start a second task before the first has finished.
Real-time speech provides a low-friction interface for this
supervision, but it replaces orderly text turns with overlap,
backchannels, asides, and changes of mind.
Continuous interaction and complex task completion are usually
handled by different model families. A full-duplex interaction model
keeps listening and speaking on the live timeline, supporting
turn-taking, barge-in, and
backchannels~\citep{defossez2024moshi,wang2024freeze}. A long search,
difficult reasoning pass, or repository edit, however, requires
deliberate work beyond the conversation. Tool-using and coding agents
can perform that work~\citep{yao2023react}, but a sequential agent
cannot remain attentive to the live channel while it runs.

Recent systems bridge this division by keeping the interaction model
on the live channel and assigning demanding work to a delegation pool
through asynchronous handoff~\citep{qwenaudioagent,venus2026}. This
combines continuous interaction with the reasoning ability of LLMs and
the long-horizon execution of coding agents, but does not determine
how an active task should be updated, cancelled, or returned. Suppose
a user requests last night's match result, then says ``actually, the
women's match'' while the first search is running, and begins another
utterance just as the original result returns. The system must revise
the active task, reject the obsolete result, and wait for a valid
speaking opportunity. A fixed coordinator may instead launch a second
independent search, speak the first result over the user, or later
cancel the wrong task. Existing systems implement useful instances
of handoff, acknowledgement, and delivery control~\citep{qwenaudioagent,venus2026,gander2026,lin2025asyncvoice},
but these decisions are generally coupled in system-specific
heuristics. When the same failure recurs, a trace may not clearly
identify whether delegation, task interpretation, background
execution, or delivery should change. A bounded repair is then
difficult to isolate, evaluate, and transfer.

We present \emph{DuplexAgent}, a full-duplex collaboration system
whose harness makes this control explicit in six editable modules:
Policy decides whether a turn is answered, delegated, or treated as
task control; Intake interprets the request and the live task it
concerns; Hold maintains the conversation while work is pending;
Worker selects the delegate and the result contract; Deliver controls
when and how a result is returned; and Memory preserves the context
shared by the conversation and the background work. The interaction
model remains the conversational entry, while a \emph{delegation pool}
of reasoning LLMs and coding agents carries delegated work. A common
harness coordinates their exchange through model-specific adapters,
supporting both open-weight and hosted interaction models and one or
several delegates.

Separating these modules gives each failure a place to be
investigated and provides the basis for \emph{Duplex-Harness-RSI}.
In the match-query example, the trace can distinguish a task that was
not updated from an obsolete result that was delivered. The diagnosis
points to Intake, Deliver, or a repair involving both, rather than a
change to the whole harness. The proposed change is then specific
enough to evaluate against the failed conversation and other tests.

Each test is a timed conversation assembled from a request type and
an interaction pattern. It states the expected behaviour, and the
trace attributes a failure to the responsible module. A reasoning LLM
from the delegation pool acts as the \emph{Exam Planner}: it reads
attributed failures and the repair archive, then adjusts the next
exam. The simulator generates the conversations. A coding agent from
the same pool acts as the \emph{Harness Editor} and proposes changes
to the modules named by the diagnosis. Evaluation compares each
candidate with the current harness on paired runs and checks a disjoint
protection set before retention.

\begin{figure}[t]
\centering
\includegraphics[width=\linewidth]{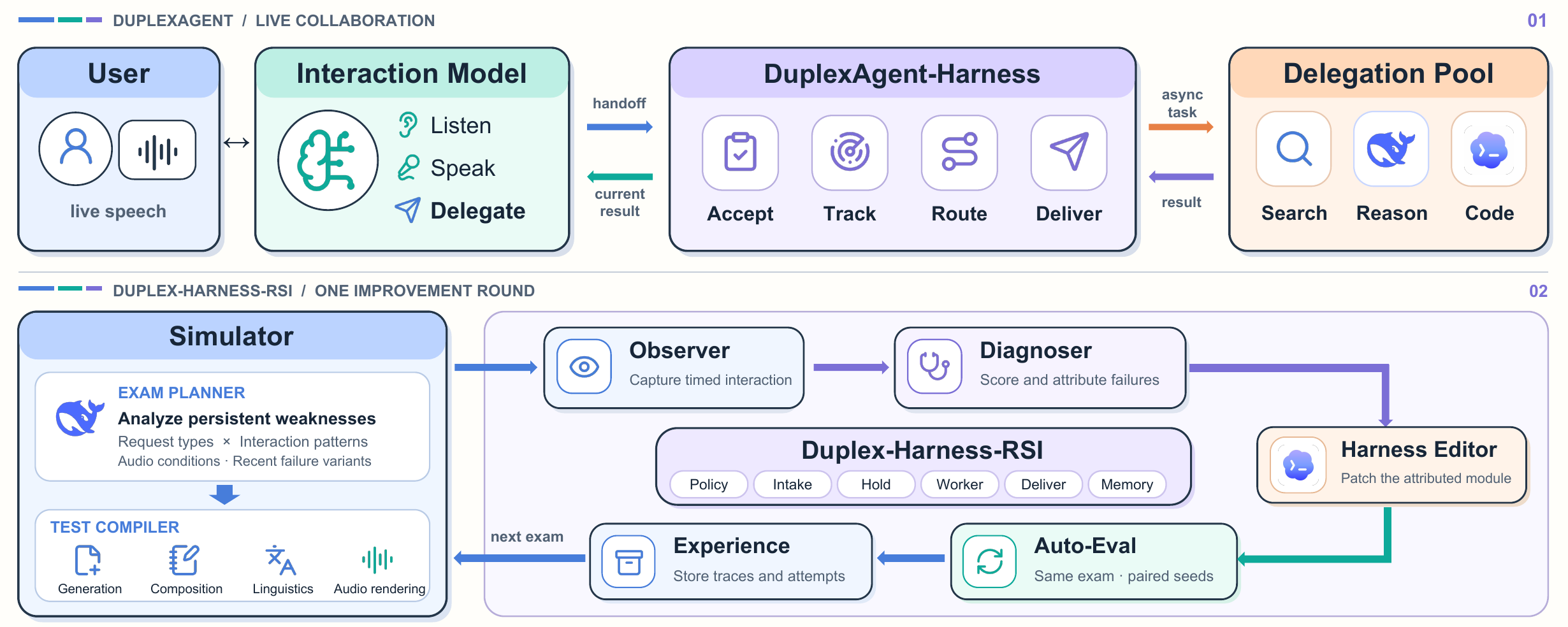}
\caption{Schematic of DuplexAgent and Duplex-Harness-RSI.}
\label{fig:overview}
\end{figure}

Section~\ref{sec:related} reviews related work.
Section~\ref{sec:method} presents the six modules, test generation,
failure attribution, and the improvement loop.
Section~\ref{sec:impl} describes the session engine, the adapters,
and the RSI configuration used in the experiments.
Section~\ref{sec:experiments} reports spoken knowledge, tool use,
duplex timing, and harness improvement with two interaction models.
Figure~\ref{fig:overview} summarises the system and improvement loop.

\setlength{\textfloatsep}{8pt plus 2pt minus 2pt}
\paragraph{Contributions.}
\begin{enumerate}\setlength{\itemsep}{4pt}
\item \emph{DuplexAgent.}
A complete collaboration system that attaches a range of
full-duplex interaction models as the conversational entry and a
range of reasoning LLMs and coding agents as the delegation pool,
and that expresses the collaboration workflow as six editable
harness modules.
\filbreak
\item \emph{Duplex-Harness-RSI.}
A recursive improvement loop that reuses the delegation pool. A
reasoning LLM acts as Exam Planner, selecting tests from attributed
failures and the repair archive. A coding agent acts as Harness Editor,
revising the relevant modules. Simulation and verifiable evaluation
connect these revisions to observed outcomes, so the capabilities
that serve the user also improve the harness.

\item \emph{A reusable collaboration simulator.}
An automatic pipeline for content generation, temporal composition,
speech rendering, and evaluation, with coverage organized by nineteen
request types and twenty-two interaction patterns. It supplies the
tests and failure traces used by Duplex-Harness-RSI and can generate
training or evaluation conversations for other interaction systems.

\end{enumerate}

\section{Related work}
\label{sec:related}
\begingroup
\setlength{\parskip}{0.28pc}
\makeatletter
\def\paragraph{\@startsection{paragraph}{4}{\z@}{0.7ex plus
0.2ex minus .1ex}{-1em}{\normalsize\bf}}
\makeatother

\paragraph{Full-duplex models and delegated agents.}
Full-duplex speech research has explored both end-to-end and modular routes. Models such as Moshi~\citep{defossez2024moshi}, Freeze-Omni~\citep{wang2024freeze}, SALMONN-omni~\citep{yu2024salmonn}, and MiniCPM-o~4.5~\citep{cui2026minicpmo} address streaming speech understanding and generation, while modular systems study controllable components, acoustic-semantic separation, and voice or role conditioning~\citep{liu2025x,liu2026hierarchical,roy2026personaplex,chen2025fireredchat}. A second line of work connects the live channel to background computation: MoshiRAG~\citep{chien2026moshirag} adds asynchronous retrieval, Qwen-Audio-Agent~\citep{qwenaudioagent} delegates background work, Realtime-Venus~\citep{venus2026} coordinates asynchronous returns, and Gander~\citep{gander2026} connects interaction to long-horizon execution. Recent systems and benchmarks also emphasize multi-round state and realistic overlap rather than isolated turns~\citep{lin2025fullv15,he2026mtr,wang2026full}. DuplexAgent builds on this architectural direction and studies how its coordination mechanisms can be diagnosed and revised from experience.

\paragraph{Evaluation of spoken agents.}
VoiceBench~\citep{chen2024voicebench} evaluates voice assistants across spoken tasks and input conditions. Full-Duplex-Bench~\citep{lin2025full} measures turn-taking, with later versions extending overlap handling and tool use~\citep{lin2025fullv15,lin2026fullv3}. BFCL~\citep{patil2025bfcl} evaluates structured function calling, while $\tau$-Voice~\citep{ray2026tauvoice} studies voice agents on service tasks. These benchmarks provide external measures of knowledge, tool execution, and interaction. Our simulator complements them with explicit task transitions and module-attributed failures for harness revision. External benchmark scores and simulator collaboration scores answer different questions and are reported separately.

\paragraph{Recursive harness improvement.}
Automated agent design and harness optimization treat prompts, tools, or executable coordination as objects that can be searched and revised~\citep{hu2025automated,lee2026meta}. Self-harness and recursive improvement systems extend this idea by using traces, scores, and prior attempts to guide later changes~\citep{zhang2026self,tan2026metarsi}. Related work also explores modular repair and training-time optimization of agent workflows~\citep{wu2026modularrsi,luo2025agent}. Duplex-Harness-RSI applies harness revision to live spoken collaboration, where correctness includes task identity, conversational continuity, and the timing of delivery. Its Exam Planner adapts the test distribution, and its Harness Editor changes the harness modules under a fixed evaluation boundary.
\endgroup
\setlength{\textfloatsep}{20pt plus 2pt minus 4pt}
\section{Method}
\label{sec:method}
\FloatBarrier

A deployed DuplexAgent system is
\begin{equation}
S=(F,B,H),
\label{eq:system}
\end{equation}
where $F$ is the full-duplex interaction model, $B$ is the delegation pool, and $H$ is the collaboration harness. This decomposition separates the models that produce an answer from the harness that decides how that answer enters the live conversation. $F$ listens and speaks on the user-facing channel, members of $B$ carry out longer jobs asynchronously, and $H$ maintains the identity, timing, and context of the work between them. Duplex-Harness-RSI improves $H$ while keeping the attached models fixed. Its Exam Planner and Harness Editor are reasoning and coding roles drawn from the same pool $B$ that serves user requests, so the improvement loop is part of the deployed system.

\subsection{The collaboration harness}
\label{sec:harness}

\begin{figure}[H]
\centering
\includegraphics[width=\linewidth]{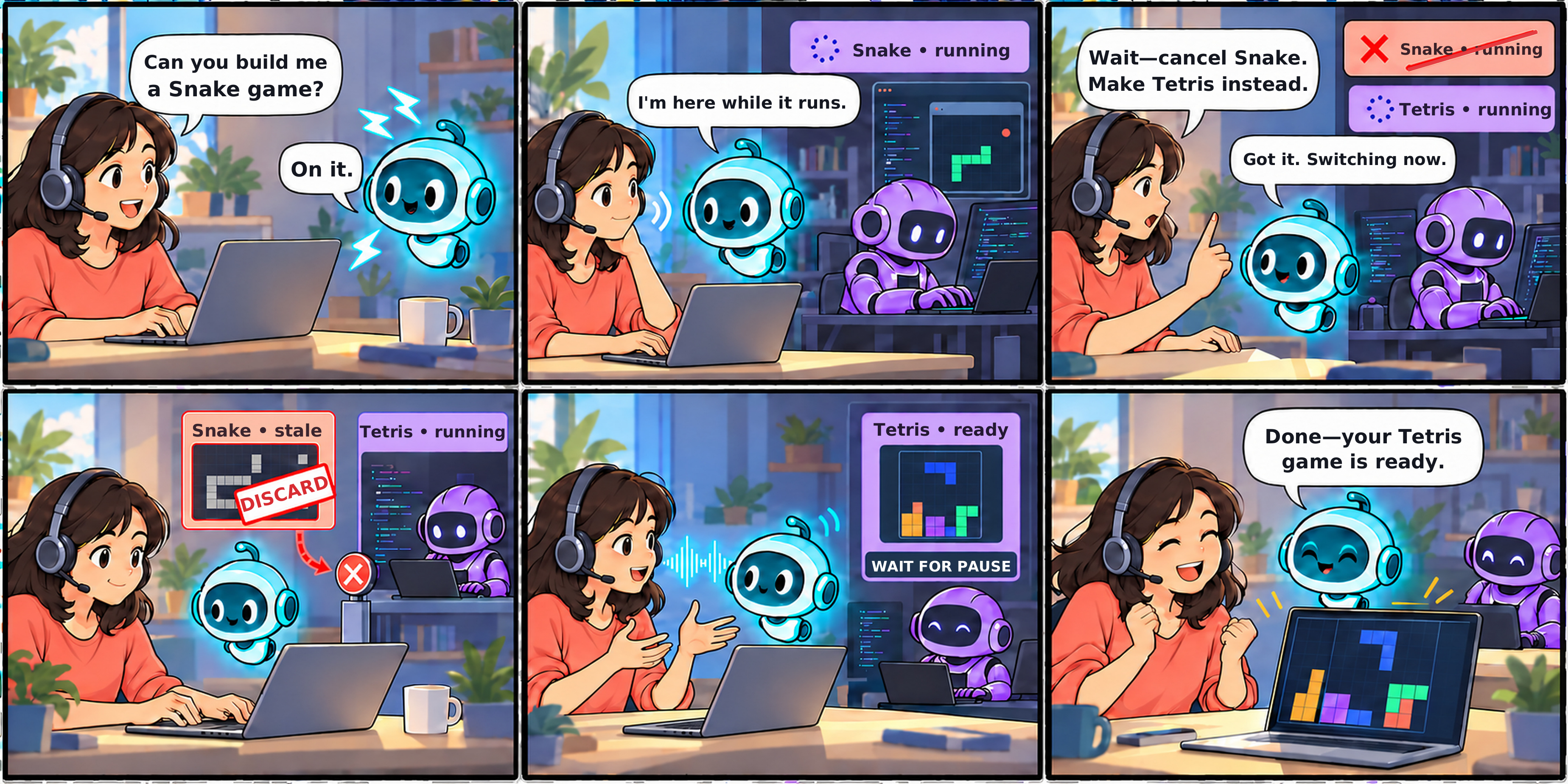}
\caption{A full-duplex collaboration session. The user requests Snake, then cancels it and asks for Tetris while work is running. The obsolete Snake result is discarded. The current Tetris result returns through the interaction model after the user pauses.}
\label{fig:comic}
\end{figure}

The harness exists because a delegated result is useful only when it still answers the user's current request and can be delivered without breaking the conversation. Figure~\ref{fig:comic} makes this requirement concrete: the user changes the game while the first task is running, so the system must update the active goal, retire the obsolete work, and wait for an appropriate moment to speak.

The harness comprises
\begin{equation}
H=(\mathrm{Policy},\mathrm{Intake},\mathrm{Hold},
\mathrm{Worker},\mathrm{Deliver},\mathrm{Memory}).
\label{eq:harness}
\end{equation}
Table~\ref{tab:modules} defines the decision owned by each module and the surface on which it can be changed. Policy chooses the route: answer, delegate, or task control. Intake identifies the operation and the live task it concerns. Hold and Deliver protect the conversational rhythm. Worker governs the handoff to a delegate, and Memory keeps the relevant context available to both the live conversation and the background work. Each module has a clear input, decision, and observable effect, so a failure can be attributed to a bounded part of the harness and a later revision can be checked against the same behaviour. An interaction-model adapter carries these decisions to the attached model without changing their meaning.

\begin{table}[ht]
\centering
\small
\caption{Six editable modules of the DuplexAgent harness.}
\label{tab:modules}
\setlength{\tabcolsep}{5pt}
\renewcommand{\arraystretch}{1.2}
\arrayrulecolor{ResultInk}
\begin{tabularx}{\linewidth}{@{}l L L@{}}
\toprule
\rowcolor{ResultHeader}Module & Collaboration decision & Editable surface \\
\midrule
\textbf{Policy} & Answer, delegate, or task control & Instructions, delegation schema, routing examples \\
\rowcolor{ResultStripe}\textbf{Intake} & Operation expressed, and which live task & Parsing, control cues, task relations, resolution rules \\
\textbf{Hold} & What the user hears while a task is pending & Receipts, progress timing, control replies \\
\rowcolor{ResultStripe}\textbf{Worker} & Delegate, context, and result contract & Routing, task instructions, result schemas, executable overrides \\
\textbf{Deliver} & When and how a current result returns & Speaking window, waiting, success or failure rendering \\
\rowcolor{ResultStripe}\textbf{Memory} & Context shared by conversation and background work & Turn assembly, retention, compaction, reset recovery \\
\bottomrule
\end{tabularx}
\end{table}

The six modules are supported by two pieces of shared state. A session engine keeps a task ledger with the task identity, version, status, and delivery history. A task can be accepted, run, completed, delivered, cancelled, or replaced; a late delegate return is still recorded, but it cannot make an obsolete task current again. Memory holds the conversational context needed to interpret the latest request and to give a delegate enough information to act. The adapter maps the harness decisions to the attached interaction model and delegate interfaces. Thus task control has a precise effect: it changes the ledger, and the ledger determines which background work remains eligible for delivery.

The Snake-to-Tetris session in Figure~\ref{fig:comic} follows one task through this protocol. Policy decides that implementation should be delegated and Memory retains the goal; Intake opens the task, Hold acknowledges it, and Worker sends the request and relevant context to a coding agent. When the user changes the goal, Intake resolves the cancellation and replacement against the ledger, Memory updates the context, and Hold acknowledges the change. A late Snake result is recorded but cannot be delivered. When Tetris becomes ready, Deliver waits for a speaking window and returns the current result through the interaction model. The six modules therefore cooperate over time, rather than acting as independent stages.

\subsection{Composing test conversations}
\label{sec:simulation}

The simulator turns the harness contract into executable conversations. It separates what the user wants from what happens while that request is being handled. A \emph{request type} specifies the requested capability and expected initial route; an \emph{interaction pattern} specifies events such as refinement, cancellation, overlap, or delayed completion. Linguistic and acoustic variation are added after these semantic expectations are fixed. The same control operation can therefore be tested across different capabilities while preserving the behaviour that the evaluator expects.

The bank contains nineteen request types. For exposition, we group their content into \emph{Conversational} requests such as social exchange and preferences; \emph{Information} requests such as factual or live-information queries; \emph{Reasoning} requests such as calculation and structured explanation; and \emph{Action} requests such as tool use and long-horizon execution. These groups describe content, not a blanket delegation rule. A familiar fact may be answered directly, a casual request may need retrieval, and an ambiguous or unsupported request may permit either a direct response or delegation. Each item records its allowed route: direct, delegate, or either.

The twenty-two patterns cover \emph{Nominal} completion; \emph{Task control} including refinement, replacement, cancellation, and status; \emph{Concurrent work} with new requests before existing work finishes; \emph{Full-duplex overlap} including barge-in, result arrival during user speech, and backchannels; and \emph{Continuity and recovery} including delay, failure, follow-up, and reset. Only compatible request--pattern pairs are composed. The grid organizes coverage without requiring every cell to be instantiated.

With a compatible pair chosen, the pipeline first generates a request, reference answer when one is available, and natural follow-up or control turns. Quality checks test semantic fit, naturalness, answer consistency, and duplication; arithmetic references can be computed directly. Items are assigned to development, protection, and sealed sets before temporal composition, so alternate forms of one case cannot cross an evaluation boundary. A fixed core of development cases is repeated in every exam so that later rounds can be compared on the same cases. A deterministic composer then places the turns on a timeline, specifies delegate delays or failures, records the required task transitions, and defines the speaking windows in which a result is admissible. The resulting expectations are executable: answer checks may use exact, contains, regular-expression, option-letter, or semantic criteria, while tasks without a stable answer are judged by their interaction behaviour.

\begin{figure}[t]
\centering
\includegraphics[width=\linewidth]{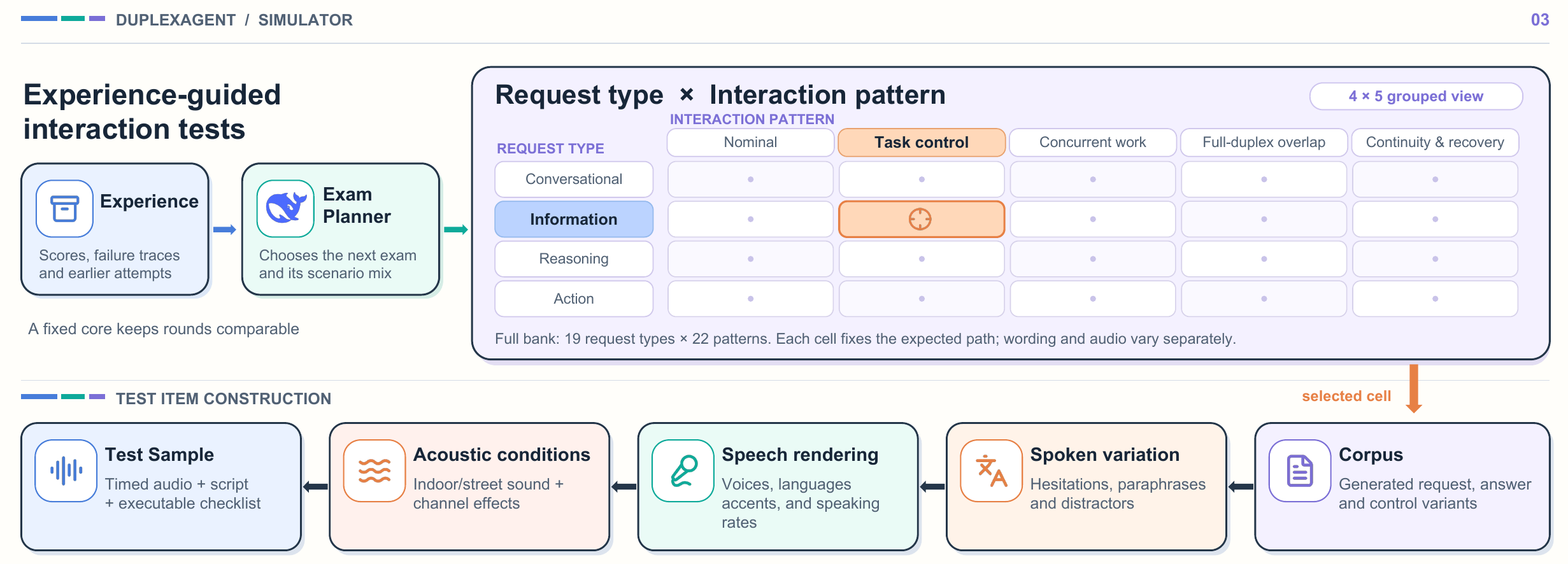}
\caption{Test-conversation construction. The Exam Planner selects coverage over request types, interaction patterns, and input conditions. Content generation and deterministic temporal composition produce a corpus with explicit expectations. Speech rendering and acoustic augmentation turn it into timed audio input.}
\label{fig:simulator}
\end{figure}

Only after these expectations are fixed do we vary the spoken surface. Hesitations, paraphrases, distractors, spoken numbers, and spelled identifiers change how a request sounds without changing what the system must do. Speech rendering and acoustic augmentation then synthesize the turns and add ambient noise or channel effects. Speaker, style, language, and pace can vary across the resulting audio, while clean and augmented versions retain the same semantics. Live evaluation streams each waveform on a realtime clock.

For the running example, an Action request is combined with replacement and result-arrival overlap. The composer schedules the goal change before Snake completes and requires only Tetris to be returned. Input variation may make cancellation hesitant or acoustically degraded without changing the requirement. References stay in the evaluator and never enter the interaction model's prompt.

\subsection{Trace-based diagnosis and collaboration score}
\label{sec:attribution}

A simulator case is useful for Duplex-Harness-RSI only when a failed run can be explained. In Figure~\ref{fig:overview}, the Observer records the behaviour of a run as a trace: speech activity, transcripts, delegation signals, module decisions, delegate returns, task transitions, and delivery events. The Diagnoser aligns that trace with the expected behaviour and assigns the failure signature. It also groups related outcomes and selects representative traces for the Exam Planner and the Harness Editor.

Each failed run receives a failure signature
\begin{equation}
\varphi(\tau)=(\mathrm{cause},\mathrm{module},\mathrm{pattern}).
\label{eq:signature}
\end{equation}
The Diagnoser first identifies failure flags and then selects one terminal cause in a fixed priority order. A default mapping assigns that cause to a module, and intermediate decisions and returned payloads can revise the assignment. A wrong spoken answer belongs to Deliver when the delegate payload was correct, and a missing result belongs to Worker when execution failed. Several flags may be present on one run; the terminal cause is the target that later rounds compare.

\begin{table}[b]
\centering
\small
\caption{Default failure attribution, refined using intermediate decisions and returned payloads.}
\label{tab:failures}
\setlength{\tabcolsep}{5pt}
\renewcommand{\arraystretch}{1.2}
\arrayrulecolor{ResultInk}
\begin{tabularx}{\linewidth}{@{}l l L@{}}
\toprule
\rowcolor{ResultHeader}Failure family & Default attribution & Trace evidence \\
\midrule
Perception & Outside harness & Requested turn not recovered well enough for the harness to act \\
\rowcolor{ResultStripe}Routing & Policy & Missed or extra delegation, or a reply to non-directed speech \\
Interpretation & Intake & Malformed request, or control applied to the wrong task \\
\rowcolor{ResultStripe}Continuity & Hold & Receipts conflict with speech, handoffs repeat, or pending work lacks progress \\
Execution & Worker & Execution fails, content is wrong, or the result contract is violated \\
\rowcolor{ResultStripe}Delivery & Deliver & Missing, stale, duplicated, late, unspoken, spoken over the user, or failure shown as success \\
Context & Memory & Question content or established context lost on delegation, follow-up, or reset \\
\bottomrule
\end{tabularx}
\end{table}

In a failed Snake-to-Tetris session, an obsolete Snake result may be spoken during the user's new request. The trace distinguishes incorrect task-control resolution from delivery of a result already marked obsolete. This determines whether Intake, Deliver, or a coupled repair should change; the audible error alone is insufficient.

The collaboration score summarizes these attributions. A conversation succeeds when it satisfies the item's expected behaviour. Runs in which the interaction model fails to perceive the input are excluded, because that failure is outside the harness revision. We report
\begin{equation}
Q=\text{percentage of successes, excluding perception failures}.
\label{eq:score}
\end{equation}
We compute $Q$ on the fixed core and on the protection set, which keep the same cases across rounds. $Q$ measures whether the coordination is correct; the failure signature and the trace explain the score.

\subsection{Duplex-Harness-RSI}
\label{sec:rsi}

Duplex-Harness-RSI uses the diagnosis from one exam to decide what the next exam should reveal. The Exam Planner reads attributed failures and the repair archive, then shifts emphasis among request types, interaction patterns, and input conditions. It keeps the fixed core and representative coverage, and it bounds how far the sampling weights may move. Each exam adds a weighted development sample and may include a limited number of fresh variants derived from recent failures. These variants are archived with the round rather than added to the bank permanently. Cases that retest an accepted repair stay in the plan, and repeatedly unsuccessful targets do not dominate the next exam. The planner changes what is tested, and leaves the scoring rules and the protection set as they are.

The planned exam is first run with the incumbent, the harness version currently in use. Its trace, score, failure signature, representative cases, and the visible part of the repair archive form the evidence given to the Harness Editor. The editor must turn that evidence into a testable hypothesis: it identifies the decision that produced the failure, proposes a coherent change on the declared editable surfaces, and records what the change is expected to improve. The simulator and the evaluator stay fixed, so the candidate is judged by the behaviour it changes. The excerpts below state each role's information and obligations, taken from the released contracts.

\begin{promptbox}{Exam Planner: select the next exam}
\textbf{Role.} Choose what the next exam should test.

\textbf{Inputs.} Read the request-type and interaction-pattern taxonomy,
available test inventory, per-cell outcomes and attributed failures,
the previous exam plan and its results, and the history of accepted
and rejected repairs.

\textbf{Tasks.} Assess whether the previous emphasis exposed useful
failures. Set sampling weights over request types, interaction
patterns, and input conditions. Allocate fresh variants to recent
failures the harness can address, and add cases that protect newly
repaired behaviours.

\textbf{Requirements.} Preserve the fixed core, maintain representative
coverage, and respect the limit on added cases. Use valid taxonomy
combinations. Rotate away from repeatedly unsuccessful targets;
do not change scoring rules or the protection set.

\textbf{Output.} Return one structured exam plan containing the sampling emphasis,
coverage constraints, any added cases, the previous evaluation summary,
and a short rationale for the allocation.
\end{promptbox}

\begin{promptbox}{Harness Editor: propose a targeted repair}
\textbf{Role.} Improve the collaboration harness from the supplied
evidence, using only its declared editable surfaces.

\textbf{Inputs.} Read the incumbent harness, module files and adapter
contracts, current metrics, actionable failure signatures with
representative traces, and visible prior edits, rationales, and
evaluation verdicts.

\textbf{Tasks.} Locate the decision that produced each failure and
group related cases by mechanism. Check which changes have already
been tried. State a testable hypothesis and create candidate harness
copies with coherent repairs, possibly spanning several modules.
Use visible evaluation feedback to refine rejected proposals.

\textbf{Requirements.} Preserve task validity and continuous
interaction. Edit only declared surfaces in candidate copies; keep
the incumbent, evaluator, and simulator unchanged. Do not inspect
the protection set or insert scenario-specific answers.

\textbf{Output.} For each candidate, return the revised files and a
\texttt{rationale.json} describing the primary module, failure
analysis, hypothesis, change summary, expected gain, risks, and
addressed signatures. Fixed evaluation determines retention.
\end{promptbox}

The evaluator checks six invariants: delegation originates from the interaction model; the harness does not mute or interrupt its speech; acceptance precedes completion; only a current result may be delivered, at most once; results return through the interaction model; and test reference answers do not enter its prompt. The session engine, adapter contracts, simulator, and scoring rules are outside the editor's write surfaces. These constraints keep the meaning of a successful revision fixed while the harness modules change.

The incumbent and the candidate are evaluated on the same exam. A \emph{win} is an incumbent failure that the candidate makes acceptable; a \emph{loss} is the reverse. A run that fails only because the interaction model did not perceive the input is not counted as a harness loss. This is the same exclusion used in $Q$.

The development gate requires more wins than losses, preserved invariants, and cause regressions within a fixed tolerance. It reports which signatures changed, so the editor can compare its hypothesis with the measured effect. Candidates are ranked by their paired wins and losses. A candidate that passes is checked on the protection set in separate executions. The protection gate requires at least as many wins as losses and preserved invariants. Protection traces stay hidden from the editor. A rejected candidate does not replace the incumbent.

The repair archive keeps exam plans, traces, candidate changes, rationales, and visible verdicts for both accepted and rejected attempts. In Figure~\ref{fig:overview}, experience is this archive carried into the next round. The Exam Planner and the Harness Editor use it to avoid repeating unsuccessful changes and to build on repairs that have already helped.

Let $g_0$ denote the campaign's initial harness and $g_k$ its $k$th retained version. A round checkpoint $r_t$ reports the incumbent after $t$ completed improvement rounds. Rejection keeps the current version, and a repair that touches several modules still creates one retained version.

The loop plans an exam, runs and diagnoses it, edits the harness, applies the development and protection gates, and writes the repair archive. The reasoning and coding capabilities come from the delegation pool already used by DuplexAgent. The interfaces, test expectations, and gates stay fixed while that pool revises the harness.
\FloatBarrier

\section{Implementation}
\label{sec:impl}

The session engine consumes events for user and agent speech, partial and final transcripts, and delegation or control signals. It sends instructions, tool definitions, context updates, and returned results through an adapter. Each adapter declares its capabilities, including native function calling, textual delegation markers, context injection, and the treatment of interim tool outputs. The harness uses these declarations to choose a compatible receipt and delivery path. This separates harness decisions from the transport details of an open-weight or hosted realtime interaction model.

Worker selects a delegate according to the requested capability and supplies the current goal, relevant context, and a result contract. In the reported setup, DeepSeek-v4-flash handles reasoning and general delegated work, while Codex supplies coding capability. During Duplex-Harness-RSI, DeepSeek-v4-flash is reused as the Exam Planner and Codex as the Harness Editor. The planner runs offline, and the editor runs in an isolated workspace that contains the incumbent harness and the visible part of the repair archive.

A harness version consists of a manifest and its declared module files, including instructions, configuration, and executable overrides. The evaluator derives a patch from the editor's filesystem changes and rejects undeclared paths. The editor cannot change scoring rules, import the simulator into the harness, or insert scenario-specific answers.

The experiments use the default configuration of the released implementation, summarized in Table~\ref{tab:rsi-config}. Each round runs the live interaction model and the delegation pool. The six harness modules are revised, and the model weights stay fixed.

\begin{table}[ht]
\centering
\small
\caption{Configuration used in the RSI experiments.}
\label{tab:rsi-config}
\setlength{\tabcolsep}{5pt}
\renewcommand{\arraystretch}{1.2}
\arrayrulecolor{ResultInk}
\begin{tabularx}{\linewidth}{@{}l L@{}}
\toprule
\rowcolor{ResultHeader}Component & Experimental setting \\
\midrule
\textbf{Content bank} & Content split before temporal composition into development, protection, and sealed sets (60/20/20). \\
\rowcolor{ResultStripe}\textbf{Exam Planner} & DeepSeek-v4-flash; bounded weights over request types, interaction patterns, and input conditions, with request-coverage floors. \\
\textbf{Adaptive exam} & Fixed core share 40\%; fresh-item budget of 30 per round, with weighted sampling of the remaining cases. \\
\rowcolor{ResultStripe}\textbf{Harness Editor} & Codex using \texttt{gpt-5.6-sol}; three candidates and up to two refinement attempts per round. \\
\textbf{Development gate} & More paired wins than losses, preserved collaboration invariants, and cause regressions within a fixed tolerance. \\
\rowcolor{ResultStripe}\textbf{Protection gate} & A separate protection set; at least as many paired wins as losses and preserved invariants. \\
\textbf{Repair archive} & Attributed outcomes, candidate changes, and visible verdicts for accepted and rejected attempts. \\
\bottomrule
\end{tabularx}
\end{table}

\section{Experiments}
\label{sec:experiments}

We evaluate the complete system on external tasks and examine harness revision separately on simulator conversations. External comparisons measure the combined effect of delegation and the harness. The ablation then compares ways of revising that harness. Percentage differences below are stated in percentage points.

\subsection{Evaluation setup}
\label{sec:eval-setup}

\paragraph{Spoken knowledge.}
We use OpenBookQA (OBQA) and Massive Multi-task Spoken Understanding (MMSU) from VoiceBench~\citep{chen2024voicebench}. Both present spoken multiple-choice questions. OBQA emphasizes elementary science, whereas MMSU spans a wider range of subjects. The response is parsed for an option letter and compared with the reference.

\paragraph{Executable tool use.}
The BFCL evaluation~\citep{patil2025bfcl} uses spoken requests from five single-turn subsets in the v3 release: simple, multiple, parallel, parallel-multiple, and irrelevance. The first four vary the number of available functions and required calls; irrelevance tests whether the system abstains from calling an unsuitable function. Calls are scored against the benchmark's function and argument requirements. We report each subset and their unweighted mean.

FDB-v3~\citep{lin2026fullv3} evaluates tool use under naturally disfluent speech. Selection F1 measures agreement between expected and selected calls. Argument accuracy uses strict character-level exact-string matching, and Pass@1 requires the correct tool set with every argument matching exactly.

\paragraph{Duplex timing.}
We use FDB-v1~\citep{lin2025full} pause handling, turn taking, and user interruption. Input is streamed on a realtime clock; output speech is aligned to the evaluation windows. Takeover rate (TOR) is the fraction of samples with a substantive response rather than silence or a brief backchannel. Lower pause TOR means fewer premature responses during a hesitation. Higher turn TOR and interrupt TOR mean the system responds more often after a completed turn or an interruption. We report the synthetic and Candor pause conditions separately.

\paragraph{Systems and comparison scope.}
Tables~\ref{tab:agentic} and \ref{tab:intel-duplex} include standalone interaction models and systems that add asynchronous delegation. MoshiRAG adds retrieval to Moshi. Gander connects an interaction model to long-horizon execution. Realtime-Venus-Omni is the Realtime-Venus system attached to Realtime-Omni-9B. Qwen-Audio-Agent and DuplexAgent are each run on VoiceChat and on Qwen-Realtime. Throughout the tables and figures, VoiceChat denotes NVIDIA-NemotronLabs-VoiceChat-11B~\citep{nvidiavoicechat} and Qwen-Realtime denotes Qwen-Audio-3.0-Realtime. In the Qwen-Realtime block, Qwen-Audio-Agent and DuplexAgent delegate reasoning to DeepSeek-v4-flash. Comparisons within a block hold the interaction model fixed. Comparisons across blocks also change the interaction model.

\paragraph{Simulator and improvement setup.}
DeepSeek-v4-flash generates test content and serves as the Exam Planner; Codex serves as the Harness Editor. Qwen3-TTS renders the composed conversations as speech, varying speaker, language, and speaking style, and acoustic augmentation adds noise and channel effects. The simulator follows Section~\ref{sec:simulation}. External benchmarks keep their own inputs.

VoiceChat and Qwen-Realtime are revised in separate campaigns with the configuration in Table~\ref{tab:rsi-config}. Model weights stay fixed. The fixed core is repeated in every exam so that collaboration can be compared across revisions. The exams used for editing also include adaptively chosen development cases and fresh variants.

\subsection{External task performance}
\subsubsection{Executable tool use}

\begin{table}[ht]
\centering
\caption{Executable tool use on BFCL and FDB-v3. DuplexAgent scores higher than the standalone interaction model and Qwen-Audio-Agent on both attachments. Scores are percentages. $^*$ denotes VoiceChat and $^\dagger$ denotes Qwen-Realtime.}
\label{tab:agentic}
\setlength{\tabcolsep}{3.35pt}
\renewcommand{\arraystretch}{1.35}
\arrayrulecolor{ResultInk}
\resizebox{\linewidth}{!}{%
\begin{tabular}{@{}lc *{6}{c} *{3}{c}@{}}
\toprule
& & \multicolumn{6}{c}{BFCL-v3} & \multicolumn{3}{c}{FDB-v3} \\
\cmidrule(lr){3-8} \cmidrule(lr){9-11}
\rowcolor{ResultHeader}Method & Delegate & Average & Simp. & Mult. & Para. & P-M & Irre. & F1 & Argu. & Pass@1 \\
\midrule
Moshi & \no & 20.0 & 0.0 & 0.0 & 0.0 & 0.0 & 100.0 & 0.0 & 0.0 & 0.0 \\
\rowcolor{ResultStripe}MoshiRAG & \yes & 20.0 & 0.0 & 0.0 & 0.0 & 0.0 & 100.0 & 0.0 & 0.0 & 0.0 \\
\midrule
MiniCPM-o~4.5 & \no & 20.0 & 0.0 & 0.0 & 0.0 & 0.0 & 100.0 & 0.0 & 0.0 & 0.0 \\
\rowcolor{ResultStripe}Gander & \yes & 17.7 & 0.0 & 0.0 & 0.0 & 0.0 & 88.3 & 0.0 & 0.0 & 0.0 \\
\midrule
Realtime-Omni-9B & \no & 68.8 & 74.0 & 57.0 & 71.0 & 53.0 & 89.2 & 7.0 & 4.2 & 2.0 \\
\rowcolor{ResultStripe}Realtime-Venus-Omni & \yes & 46.5 & 48.5 & 41.0 & 35.0 & 22.0 & 85.8 & 5.1 & 3.3 & 1.0 \\
\midrule
VoiceChat & \no & 53.5 & 54.5 & 60.0 & 36.0 & 22.0 & 95.0 & 60.2 & 26.2 & 16.0 \\
\rowcolor{ResultStripe}Qwen-Audio-Agent$^*$ & \yes & 24.9 & 3.5 & 2.0 & 12.0 & 7.0 & \textbf{100.0} & 19.2 & 13.5 & 5.0 \\
\rowcolor{ResultHighlight}\textbf{DuplexAgent$^*$} & \yes & \textbf{75.8} & \textbf{77.0} & \textbf{78.0} & \textbf{77.0} & \textbf{57.0} & 90.0 & \textbf{97.4} & \textbf{57.0} & \textbf{47.0} \\
\midrule
Qwen-Realtime & \no & 82.0 & 88.0 & 86.0 & 85.0 & 66.0 & 85.0 & \textbf{87.0} & 47.8 & 39.0 \\
\rowcolor{ResultStripe}Qwen-Audio-Agent$^\dagger$ & \yes & 80.8 & 88.0 & 85.0 & 77.0 & 58.0 & 95.8 & 75.5 & 42.7 & 30.0 \\
\rowcolor{ResultHighlight}\textbf{DuplexAgent$^\dagger$} & \yes & \textbf{90.8} & \textbf{92.0} & \textbf{88.0} & \textbf{96.0} & \textbf{78.0} & \textbf{100.0} & 84.8 & \textbf{55.0} & \textbf{49.0} \\
\bottomrule
\end{tabular}%
}
\end{table}

\begin{figure}[h]
\centering
\includegraphics[width=\linewidth]{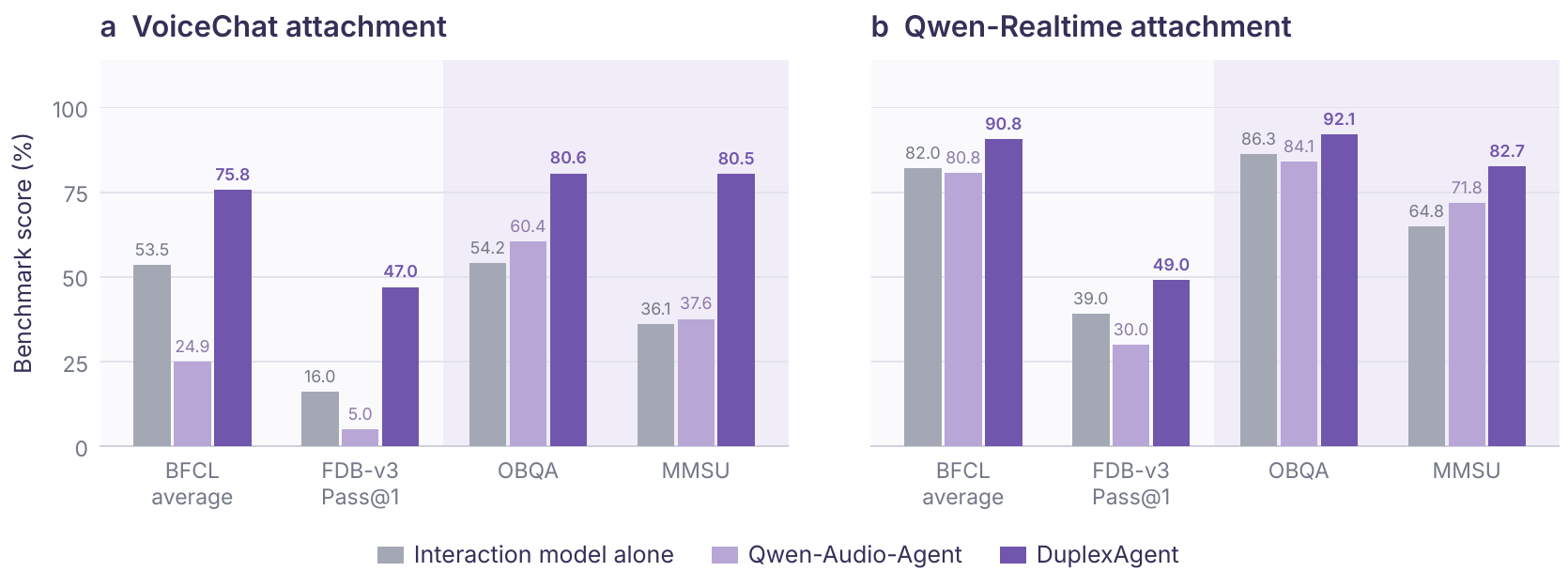}
\caption{DuplexAgent scores above both its standalone interaction model and Qwen-Audio-Agent on the same attachment.}
\label{fig:attachment-gains}
\end{figure}

\paragraph{Executable calls require a precise handoff.}
Table~\ref{tab:agentic} separates two kinds of tool behaviour. A standalone interaction model that can bind the benchmark schema keeps a usable BFCL score: Realtime-Omni-9B averages 68.8, VoiceChat 53.5, and Qwen-Realtime 82.0. Moshi and MiniCPM-o~4.5 have no such binding, so their four calling subsets are 0.0 and only irrelevance, which rewards an empty call list, reaches 100.0. Adding a background delegate does not create that binding, and where the interaction model already emits calls, the added system can replace them with a description. MoshiRAG, Gander, Realtime-Venus-Omni, and Qwen-Audio-Agent therefore leave executable accuracy at or below the matching interaction model. The drop is largest for VoiceChat and Realtime-Omni-9B, the interaction models that were already calling.

Those systems hand the delegate a description rather than a benchmark call. Gander's interaction model opens work with \texttt{task\_start}, \texttt{task\_send}, and \texttt{task\_resolve}; the argument of \texttt{task\_start} is a short task title, and a coding agent then acts with its own file and command tools. Realtime-Venus-Omni writes the request as prose inside a \texttt{<delegate>} span and dispatches that sentence. Qwen-Audio-Agent answers in speech when it can, and otherwise calls \texttt{spawn\_thinking} with a natural-language restatement of the objective. MoshiRAG retrieves passages for the spoken reply. A request such as ``implement this feature'' or ``draft this section'' can succeed under that contract, because the delegate chooses its own actions from a fuzzy objective. BFCL and FDB-v3 ask for something narrower. The scored object is the benchmark function name together with its required arguments: one call, several calls, or no call, and on FDB-v3 an exact argument string recovered from disfluent speech.

The recorded handoffs match that gap. Gander's calls are task titles, so the BFCL calling subsets and all three FDB-v3 numbers stay at 0.0, and titles emitted on items that should abstain lower irrelevance from 100.0 to 88.3. Realtime-Omni-9B already fills the function on clean BFCL speech; FDB-v3 is the harder surface, and Realtime-Omni-9B withholds the call on nearly every item (Pass@1 2.0). Realtime-Venus-Omni then substitutes a sentence for the call the interaction model had produced. BFCL average falls from 68.8 to 46.5, parallel from 71.0 to 35.0, and parallel-multiple from 53.0 to 22.0, because one prose span must be parsed back into several typed arguments. FDB-v3 Pass@1 stays at 1.0. Qwen-Audio-Agent on VoiceChat withholds calls the interaction model was making: simple accuracy falls from 54.5 to 3.5, FDB-v3 Pass@1 from 16.0 to 5.0, and irrelevance rises to 100.0. On Qwen-Realtime the interaction model already calls accurately, and the same restatement still trims parallel accuracy from 85.0 to 77.0, parallel-multiple from 66.0 to 58.0, and FDB-v3 Pass@1 from 39.0 to 30.0.

DuplexAgent keeps the decision to delegate separate from the content of the handoff. Policy leaves a direct reply on the interaction model and delegates the request that needs a tool or a harder computation. Worker receives the official schema and the arguments carried by the user turn, and Memory retains the identifiers and constraints that a title or a short restatement drops. Deliver returns only the current result, in a speaking window. Those rules are the harness after Duplex-Harness-RSI (Section~\ref{sec:eval-rsi}), scored on the same waveforms, schemas, and checks as the other rows. On VoiceChat, BFCL average rises from 53.5 to 75.8 and FDB-v3 from 60.2 / 26.2 / 16.0 to 97.4 / 57.0 / 47.0. Irrelevance falls from 95.0 to 90.0: recovering required calls also admits some unnecessary ones. On Qwen-Realtime, BFCL average rises from 82.0 to 90.8, parallel-multiple from 66.0 to 78.0, and irrelevance from 85.0 to 100.0. Selection F1 moves from 87.0 to 84.8, while argument accuracy rises from 47.8 to 55.0 and Pass@1 from 39.0 to 49.0, so the completed calls are more exact even where selection is slightly less complete.

\subsubsection{Spoken knowledge and duplex timing}

\begin{table}[ht]
\centering
\caption{Spoken knowledge and duplex timing. DuplexAgent raises spoken-knowledge accuracy on both attachments, while duplex timing stays near the attached interaction model. Marks follow Table~\ref{tab:agentic}.}
\label{tab:intel-duplex}
\setlength{\tabcolsep}{5pt}
\renewcommand{\arraystretch}{1.35}
\arrayrulecolor{ResultInk}
\resizebox{\linewidth}{!}{%
\begin{tabular}{@{}lc *{2}{c} *{4}{c}@{}}
\toprule
& & \multicolumn{2}{c}{Intelligence} & \multicolumn{4}{c}{FDB-v1} \\
\cmidrule(lr){3-4} \cmidrule(lr){5-8}
\rowcolor{ResultHeader}Method & Delegate & OBQA & MMSU
  & \splithdr{Pause}{Synthetic~$\downarrow$}
  & \splithdr{Pause}{Candor~$\downarrow$}
  & \splithdr{Turn}{TOR~$\uparrow$}
  & \splithdr{Interrupt}{TOR~$\uparrow$} \\
\midrule
Moshi & \no & 23.8 & 24.2 & 1.000 & 0.991 & 1.000 & 0.94 \\
\rowcolor{ResultStripe}MoshiRAG & \yes & 30.8 & 28.1 & 1.000 & 0.981 & 1.000 & 0.83 \\
\midrule
MiniCPM-o~4.5 & \no & 69.6 & 55.0 & 0.132 & 0.222 & 0.898 & 0.85 \\
\rowcolor{ResultStripe}Gander & \yes & 82.4 & 62.7 & 0.000 & 0.019 & 0.390 & 0.72 \\
\midrule
Realtime-Omni-9B & \no & 84.1 & 64.5 & 0.000 & 0.019 & 0.864 & 0.68 \\
\rowcolor{ResultStripe}Realtime-Venus-Omni & \yes & 41.4 & 27.4 & 0.000 & 0.019 & 0.797 & 0.78 \\
\midrule
VoiceChat & \no & 54.2 & 36.1 & 0.162 & 0.343 & \textbf{0.983} & \textbf{1.00} \\
\rowcolor{ResultStripe}Qwen-Audio-Agent$^*$ & \yes & 60.4 & 37.6 & 0.206 & 0.324 & 0.966 & 0.99 \\
\rowcolor{ResultHighlight}\textbf{DuplexAgent$^*$} & \yes & \textbf{80.6} & \textbf{80.5} & \textbf{0.117} & \textbf{0.241} & 0.961 & 0.965 \\
\midrule
Qwen-Realtime & \no & 86.3 & 64.8 & \textbf{0.074} & 0.278 & 0.898 & 0.89 \\
\rowcolor{ResultStripe}Qwen-Audio-Agent$^\dagger$ & \yes & 84.1 & 71.8 & 0.103 & \textbf{0.176} & 0.915 & 0.92 \\
\rowcolor{ResultHighlight}\textbf{DuplexAgent$^\dagger$} & \yes & \textbf{92.1} & \textbf{82.7} & 0.147 & 0.231 & \textbf{0.966} & \textbf{0.97} \\
\bottomrule
\end{tabular}%
}
\end{table}

\paragraph{The same handoff lifts a spoken choice and can drop the call.}
Table~\ref{tab:intel-duplex} is the other half of the pattern in Table~\ref{tab:agentic}. A delegate that receives a description can still write a better spoken answer. MoshiRAG raises OBQA from 23.8 to 30.8 and MMSU from 24.2 to 28.1 while its calling subsets stay at 0.0, because retrieval supplies text for speech and never a benchmark function. Gander raises OBQA from 69.6 to 82.4 and MMSU from 55.0 to 62.7: the coding agent answers the question, and the interaction model speaks that answer, while the scored tool channel still carries only a task title. Qwen-Audio-Agent on VoiceChat raises the same cells from 54.2 to 60.4 and from 36.1 to 37.6, the small gain available when the contract prefers to answer in speech. On Qwen-Realtime the same agent lowers OBQA from 86.3 to 84.1 and raises MMSU from 64.8 to 71.8. In each of these rows, spoken knowledge and executable accuracy move apart because the handoff is a description.

\begin{figure}[t]
\centering
\includegraphics[width=\linewidth]{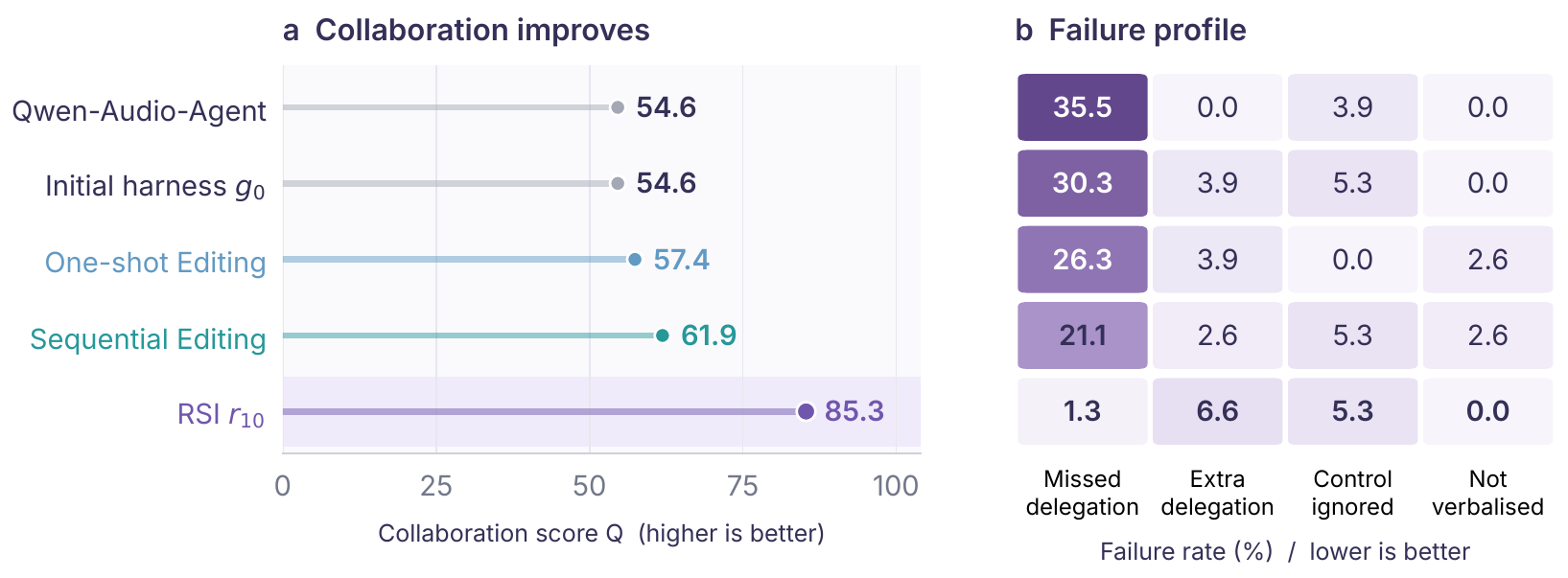}
\caption{VoiceChat harness ablation on the fixed core. The left panel is the collaboration score $Q$, and the right panel is the rate of each diagnostic failure. Duplex-Harness-RSI scores higher than one-shot or sequential editing.}
\label{fig:rsi-voicechat}
\end{figure}

Realtime-Venus-Omni shows the failure mode in which the description also displaces the answer. OBQA falls from 84.1 to 41.4 and MMSU from 64.5 to 27.4. Realtime-Omni-9B states the option letter; after the harness, the transcript that reaches the scorer is often the delegate span or a receipt, so the letter is no longer there to be parsed. Tool accuracy falls on the same attachment. A background model therefore improves a live system only when the harness decides which requests stay with the interaction model and which details travel with a delegated request.

DuplexAgent makes that separation on both attachments. A direct question remains a spoken answer; a tool request remains an executable call whose arguments were passed through. OBQA rises from 54.2 to 80.6 on VoiceChat and from 86.3 to 92.1 on Qwen-Realtime; MMSU rises from 36.1 to 80.5 and from 64.8 to 82.7. The larger gain on the weaker interaction model is the delegate supplying reasoning the interaction model does not have. The gains combine access to the delegate with the harness that selects the route, packages the handoff, and returns the result. Section~\ref{sec:eval-rsi} isolates how that harness was revised.

\paragraph{Timing remains a separate constraint.}
A description-only handoff also changes who holds the floor. Gander's pause TOR falls to 0.000 / 0.019, and its turn TOR falls from 0.898 to 0.390, so many finished user turns receive no reply while a titled task is open. Realtime-Venus-Omni keeps the Realtime-Omni-9B pause TOR and moves turn TOR from 0.864 to 0.797. Moshi and MoshiRAG speak through the user's hesitation (pause TOR at or near 1.0). DuplexAgent stays near the interaction model it is attached to. With VoiceChat, synthetic and Candor pause TOR fall from 0.162/0.343 to 0.117/0.241, while turn TOR and interrupt TOR decline slightly from 0.983/1.000 to 0.961/0.965. With Qwen-Realtime, turn TOR rises from 0.898 to 0.966 and interrupt TOR from 0.890 to 0.970. Candor pause TOR improves from 0.278 to 0.231, but synthetic pause TOR worsens from 0.074 to 0.147.

Better task performance therefore coexists with a less favorable result on one pause condition, while the live channel remains responsive.

\begin{figure}[t]
\centering
\includegraphics[width=\linewidth]{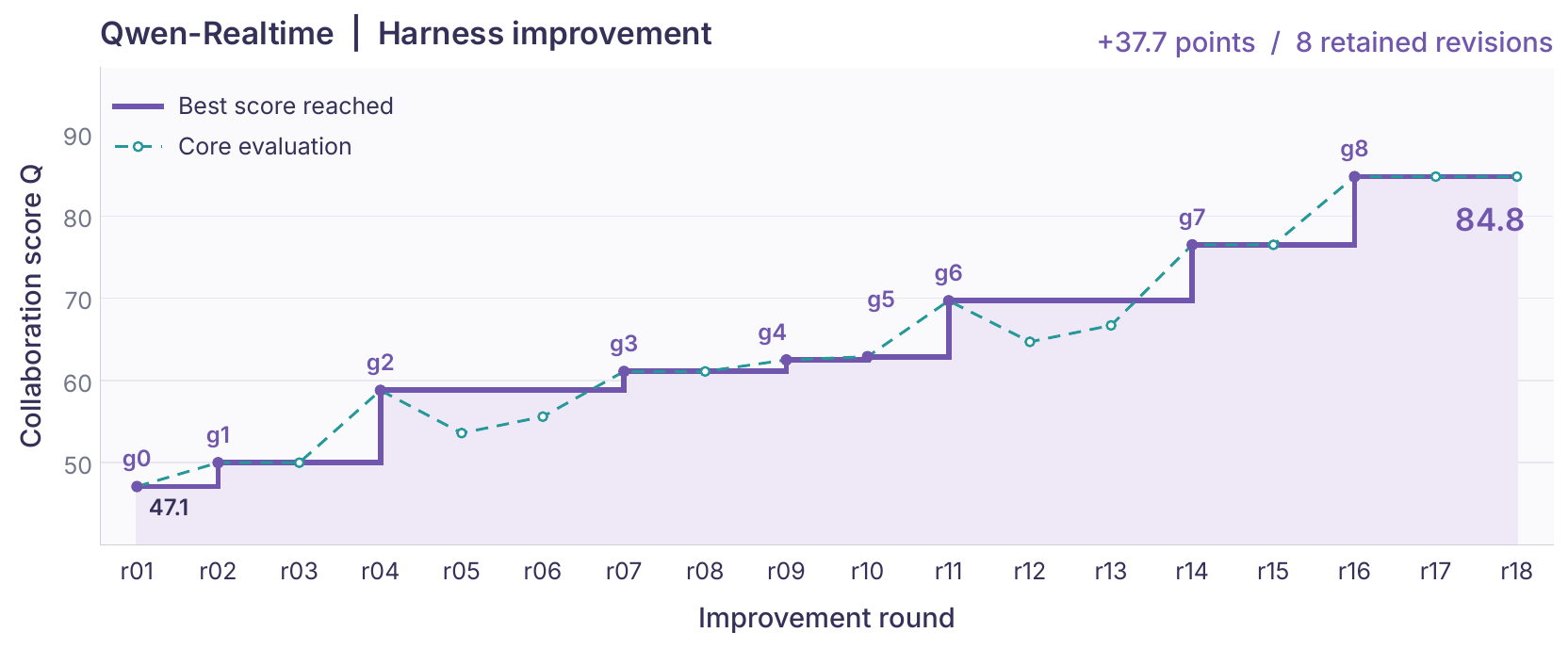}
\caption{Duplex-Harness-RSI on Qwen-Realtime. Retained versions raise the collaboration score on the fixed core from the initial harness $g_0$ to $g_8$.}
\label{fig:qwen-rsi}
\end{figure}

\subsection{Harness improvement}
\label{sec:eval-rsi}

\paragraph{Pooled editing, sequential editing, and RSI.}
We compare three ways of using the same exam material to improve the initial VoiceChat harness $g_0$:
\begin{enumerate}\setlength{\itemsep}{2pt}
\item \emph{One-shot Editing} pools all test cases encountered during the ten-round RSI campaign and gives them to the coding agent for one harness revision.
\item \emph{Sequential Editing} gives the same coding agent the exams from those ten rounds in their original order and performs ten successive revisions. It omits the per-round diagnosis and the repair archive used by Duplex-Harness-RSI.
\item \emph{Duplex-Harness-RSI} performs ten rounds. Each round plans an exam, runs and diagnoses it, edits the harness, and applies the development and protection gates. The Exam Planner and the Harness Editor use that round's diagnosis and the repair archive. We report the incumbent at checkpoint $r_{10}$.
\end{enumerate}

The exams used for editing extend beyond the fixed core. One-shot Editing receives all of them in one pool. Sequential Editing replays them in the original order for the same number of rounds. Figure~\ref{fig:rsi-voicechat} compares the resulting harnesses on the fixed core.

Qwen-Audio-Agent and $g_0$ both score 54.6, with different rates of missed and extra delegation. One-shot Editing reaches 57.4 and Sequential Editing reaches 61.9. The ten-round checkpoint $r_{10}$ reaches 85.3, improving on $g_0$ by 30.7 points and on Sequential Editing by 23.4 points. Repeated access to the same exams improves the harness. The full loop, which adds diagnosis and the repair archive, produces a substantially larger gain.

Missed delegation falls from 30.3\% of items in the fixed core under $g_0$ to 1.3\% under Duplex-Harness-RSI. Extra delegation rises from 3.9\% to 6.6\%, and ignored control stays at 5.3\%. The main gain is recovered delegation, with a remaining tradeoff in unnecessary calls and unresolved control.

\paragraph{Version lineage and progress.}
Figure~\ref{fig:qwen-rsi} follows the Qwen-Realtime campaign. It is separate from the VoiceChat checkpoint $r_{10}$: each attachment has its own initial harness and its own sequence of exams. Here the initial harness $g_0$ scores 47.1. Eight retained versions later, $g_8$ scores 84.8, a gain of 37.7 points. As defined in Section~\ref{sec:rsi}, $g_k$ counts retained versions and $r_t$ counts completed rounds, so a round can finish without producing a new $g_k$.

The dashed series records the fixed-core score after each round, including proposals that were not kept, so the line can fall. The solid staircase follows the best retained score. It steps up only when a candidate passes the development gate and the protection gate, and it stays flat when a round retains nothing, leaving the current harness in place. The late plateau is a run of rounds in which further proposals no longer passed the gates. Together with the VoiceChat ablation, the curve shows the same accumulation on a second interaction model. A repair is added only after the paired comparison supports it.

\paragraph{What the modules learned.}
Table~\ref{tab:evolution} shows the decision behind each retained repair. On Qwen-Realtime, Policy was delegating conversational turns and familiar facts too broadly. Stating the direct-answer categories explicitly, while still delegating requests that need additional capability, reduced extra calls by 50\%. On VoiceChat, Intake could treat a late cancellation as work that had already completed. Resolving it against live task state, and applying the same handling to replacement preambles, reduced ignored cancellations by 81\%. Replacement failures then fell by 65\%, and later by 83.3\%. Hold began reporting progress earlier and kept receipts brief, so missing progress updates fell by 75\%. Worker kept the requested units and precision, calculated when the request required it, and retained verified result content, so wrong answers fell by 66.7\%. Deliver was changed on both attachments: a result is spoken only after the user is silent, and an unsuccessful execution is distinguished from a successful answer. Overlaps and concealed failures both fell by 100\%. Memory is the complementary case. Delegation was already correct, but the delegate received an incomplete question or missing context. Assembling the complete turn and retaining that context reduced those failures by 60\%. These percentages come from the local exam used to test each repair. The fixed-core scores combine the repairs that the gates retained.

\begin{table}[htbp]
\centering
\small
\caption{Representative harness repairs. Each change reduces the targeted failures on the local exam used to test it. Marks follow Table~\ref{tab:agentic}.}
\label{tab:evolution}
\setlength{\tabcolsep}{5pt}
\renewcommand{\arraystretch}{1.22}
\hyphenpenalty=10000
\exhyphenpenalty=10000
\arrayrulecolor{ResultInk}
\begin{tabularx}{\linewidth}{@{}l L L L@{}}
\toprule
\rowcolor{ResultHeader}Module & Rule before repair & Rule after repair & Recorded local evidence \\
\midrule
\textbf{Policy}$^\dagger$ & Conversational and familiar-fact turns are routed too broadly. & State the direct-answer categories explicitly, while still delegating requests that need additional capability. & Extra calls reduced by 50\%. \\
\rowcolor{ResultStripe}\textbf{Intake}$^*$ & A late cancellation can be treated as work that has already completed. & Resolve the cancellation against live task state, and apply the same control handling to replacement preambles. & Ignored cancellations reduced by 81\%. Replacements reduced by 65\%, then by 83.3\%. \\
\textbf{Hold}$^\dagger$ & Pending work does not receive a timely progress update. & Report progress earlier, and keep receipts brief and compatible with the adapter. & Missing progress updates reduced by 75\%. \\
\rowcolor{ResultStripe}\textbf{Worker}$^*$ & Conversions skip the required calculation, or reformulate output that was already verified. & Preserve the requested units and precision, calculate when the request requires it, and retain verified result content. & Wrong answers reduced by 66.7\%. \\
\textbf{Deliver}$^*$ & Completion can start speech during the user's turn, and an inability to execute is reported as success. & Speak only once the user is silent, and distinguish unsuccessful execution from a successful answer. & Overlaps reduced by 100\% on both attachments. Concealed failures reduced by 100\%. \\
\rowcolor{ResultStripe}\textbf{Memory}$^*$ & Delegation is correct, but the delegate receives an incomplete question or missing context. & Assemble the complete user turn, and retain the context that a delegated request needs. & Missing-context failures reduced by 60\%. \\
\bottomrule
\end{tabularx}
\end{table}

Rejected attempts show how a broad repair can regress. A broad delivery timeout can introduce other speech conflicts, and a broad direct-answer rule can miss required delegation. The development gate therefore keeps a repair only when the hypothesis is narrow and the paired comparison supports it.

\section{Discussion and limitations}
\label{sec:discussion}

The results support a practical division of responsibility: the interaction model maintains the live conversation, the delegation pool supplies task capability, and the harness coordinates the two. The retained repairs make routing, task control, context, and the timing of delivery more precise, and spoken knowledge, executable tool use, and fixed-core collaboration rise with that precision.

These gains stop at the harness. Duplex-Harness-RSI rewrites module instructions and executable rules, and the interaction-model weights stay fixed. A repair can require silence before speech, or a complete question before delegation, while the model's perception of overlapped speech and its own judgement that a turn has ended remain as they were. When the gates accept or reject a candidate, this remainder is still in the trace. The Observer has aligned speech, delegation, and delivery with the expected behaviour, and the Diagnoser has named the cause, the module, and the pattern. The Exam Planner already turns those signatures into the next timed conversations. The record thus shows which failures a module edit can absorb, and which remain in the interaction model's own perception and timing. Agentic reinforcement learning for full-duplex collaboration continues the loop from there: the simulator keeps supplying the exams, the traces, and the failure signatures, and the update extends from the harness to the interaction model on the live timeline.

\section{Conclusion}
\label{sec:conclusion}

DuplexAgent keeps a full-duplex interaction model on the live channel and a delegation pool for reasoning and longer work, and writes their exchange as six editable modules: Policy, Intake, Hold, Worker, Deliver, and Memory. Duplex-Harness-RSI reuses the pool that serves the user. The Exam Planner chooses the next conversations from attributed failures and the repair archive, the Harness Editor revises the modules the diagnosis names, and paired checks on the fixed core and the protection set decide what is kept. The simulator makes each revision checkable, turning request types and interaction patterns into timed tests and traces. On VoiceChat and on Qwen-Realtime, the retained harness improves spoken knowledge, executable tool use, and collaboration on the fixed core, while the model weights remain those the campaign started with. The revisions show how far coordination can be improved from evidence. The same traces still contain the perception and timing that a module edit does not reach. Agentic reinforcement learning for full-duplex collaboration can take up that remainder, training the interaction model in the simulator that already evaluates the harness.

\FloatBarrier
\bibliography{refs}
\bibliographystyle{colm2024_conference}
\end{document}